\documentclass[conference]{IEEEtran}
\IEEEoverridecommandlockouts
\usepackage{cite}
\usepackage{amsmath,amssymb,amsfonts}
\usepackage{algorithmic}
\usepackage{graphicx}
\usepackage{textcomp}
\usepackage{xcolor}
\usepackage{booktabs}
\def\BibTeX{{\rm B\kern-.05em{\sc i\kern-.025em b}\kern-.08em
    T\kern-.1667em\lower.7ex\hbox{E}\kern-.125emX}}
\begin{document}

\title{Evaluating Effects of Augmented SELFIES for Molecular Understanding Using QK-LSTM
}

\author{\IEEEauthorblockN{Collin Beaudoin}
\IEEEauthorblockA{
The Pennsylvania State University\\
University Park, PA, USA \\
cpb5867@psu.edu}
\and
\IEEEauthorblockN{Swaroop Ghosh}
\IEEEauthorblockA{
The Pennsylvania State University\\
University Park, PA, USA \\}
}

\maketitle

\begin{abstract}

Identifying molecular properties, including side effects, is a critical yet time-consuming step in drug development. Failing to detect these side effects before regulatory submission can result in significant financial losses and production delays, and overlooking them during the regulatory review can lead to catastrophic consequences. This challenge presents an opportunity for innovative machine learning approaches, particularly hybrid quantum-classical models like the Quantum Kernel-Based Long Short-Term Memory (QK-LSTM) network. The QK-LSTM integrates quantum kernel functions into the classical LSTM framework, enabling the capture of complex, non-linear patterns in sequential data. By mapping input data into a high-dimensional quantum feature space, the QK-LSTM model reduces the need for large parameter sets, allowing for model compression without sacrificing accuracy in sequence-based tasks. Recent advancements have been made in the classical domain using augmented variations of the Simplified Molecular Line-Entry System (SMILES). However, to the best of our knowledge, no research has explored the impact of augmented SMILES in the quantum domain, nor the role of augmented Self-Referencing Embedded Strings (SELFIES) in either classical or hybrid quantum-classical settings. This study presents the first analysis of these approaches, providing novel insights into their potential for enhancing molecular property prediction and side effect identification. Results reveal that augmenting SELFIES yields in statistically significant improvements from SMILES by a 5.97\% improvement for the classical domain and a 5.91\% improvement for the hybrid quantum-classical domain.

\end{abstract}

\begin{IEEEkeywords}
Molecular Property Prediction, Drug Evaluation, Quantum Machine Learning, Machine Learning
\end{IEEEkeywords}

\section{Introduction}
Molecular property prediction is a cornerstone of drug discovery \cite{yang2019analyzing, wieder2020compact}. In silico methods for predicting molecular properties hold the potential to accelerate the development of safer drugs by reducing testing time and costs. By identifying molecular properties early in the development process, researchers can efficiently create novel materials with greater confidence. Moreover, detecting side effects before drug release can prevent unnecessary harm, ultimately saving lives.

Traditionally, in silico approaches relied on complex feature engineering techniques to generate molecular representations for processing \cite{lengauer2004novel, merkwirth2005automatic}. However, the limitations of these descriptor-based methods often hinder their effectiveness. The generated features are highly task-dependent, meaning they may not be transferable across different problems. Additionally, as molecular understanding evolves, previously used feature vectors may become obsolete, reducing utility.

Graph Neural Networks (GNNs) have emerged as a promising alternative, overcoming many of the limitations of traditional feature engineering. GNNs are particularly well-suited for molecular property prediction due to their ability to model molecules as graphs, a representation that naturally captures the structure of molecules. This graph-based approach removes the reliance on predefined descriptors, allowing the model to learn representations directly from the data. As a result, GNNs have shown strong performance across a range of cheminformatics tasks \cite{hu2019strategies, wu2020comprehensive}. Despite advances, GNNs face challenges, including difficulties capturing shared dependencies and scalability issues. As the size of the molecular graph increases, so does the computational complexity, with the communication cost between nodes growing exponentially. Consequently, GNNs can sometimes perform worse than other neural network architectures despite their versatile input representation \cite{mayr2018large}.

Building on the success of large language models in other fields, recent research has explored language-based approaches for molecular property prediction with promising results \cite{taylor2022galactica, m2024augmenting}. Although which model architecture is best suited for chemoinformatics \cite{jin2024large} remains an open question, language models offer several advantages that make them attractive for researchers. Notable advancements include the use of large pre-trained datasets \cite{wang2019smiles, irwin2022chemformer}, multi-modal techniques \cite{liu2024git}, data augmentation strategies \cite{tetko2019augmentation, bjerrum2017smiles}, hybrid quantum-classical machine learning \cite{beaudoin2022quantum}, and combinations of these approaches \cite{jin2024large, mao2024advancing}.

As the focus on language models continues to grow, understanding how to optimize these models is crucial. While prior studies have explored the impact of augmenting SMILES for classical networks, no research has examined the effects of augmenting SELFIES \cite{ozcceliknovo}, a language specifically designed to enhance machine understanding of chemical structures to the best of our knowledge. Moreover, no studies have investigated the potential impact of such augmentations in hybrid quantum-classical models. \emph{This study presents the first analysis of augmented SELFIES in classical and hybrid quantum-classical domains to address these gaps}.

In the following sections, we review the necessary background on MoleculeNet benchmark \cite{wu2018moleculenet}, SMILES and SELFIES formats, Long Short-Term Memory, and few of the related quantum networks (Section ~\ref{background}). We then discuss the methods used to train the models and their results (Section ~\ref{methods-results}), followed by a review of related works (Section ~\ref{related-works}). Finally, we provide concluding remarks (Section ~\ref{discussion}).

\section{Background} \label{background}

\subsection{MoleculeNet Benchmark} 

MoleculeNet is a comprehensive benchmark for evaluating machine learning models in cheminformatics \cite{wu2018moleculenet}. It curates datasets spanning quantum mechanics, physical chemistry, biophysics, and physiology. MoleculeNet establishes a preferred evaluation metric for each dataset, ensuring consistent and fair comparisons across different models. In this paper, we describe the datasets chosen to assess the performance of our model.

\subsection{Side Effect Resource (SIDER)}
A critical molecular property of drugs intended for human consumption is their associated side effects, which can significantly impact safety and efficacy. The Side Effect Resource (SIDER) dataset comprehensively collects publicly available data on known drug side effects \cite{kuhn2016sider}. This dataset compiles side effect information from various sources, offering a unified resource for understanding the adverse effects of drugs.

The SIDER dataset consists of 28 columns. The first column contains the SMILES representation of each drug molecule, providing a standardized chemical structure notation. The remaining 27 columns represent different organ classes affected by side effects, with side effects classified according to the Medical Dictionary for Regulatory Activities (MedDRA) \footnote{https://www.meddra.org/}. Each side effect is encoded with a binary value: "1" indicates that the drug is known to cause a side effect in the corresponding organ class, while "0" indicates the absence of that side effect. This structure enables systematic analysis and model training for predicting adverse drug reactions based on molecular properties.

\subsection{ROC-AUC}

The Receiver Operating Characteristic (ROC) curve is used to evaluate the performance of a binary classifier by calculating the true positive rate (TPR) against the false positive rate (FPR) across various threshold settings. The true positive rate measures the proportion of actual positives correctly identified by the model. In contrast, the false positive rate represents the proportion of negative instances incorrectly classified as positive. The ROC curve provides a comprehensive view of a model's ability to discriminate between the two classes, with the area under the curve (AUC) serving as a summary statistic that quantifies overall performance.
The ROC-AUC score, which represents the area under the ROC curve, is particularly valuable when evaluating models trained on imbalanced datasets, as it accounts for the classifier's ability to identify positive instances and avoid misclassifying negative instances correctly. This makes ROC-AUC an ideal metric for assessing models on datasets like those found in MoleculeNet, where the class imbalance is often present \cite{fawcett2006introduction}.

\subsection{Simplified Molecular-Input Line Entry System (SMILES)}

SMILES is a notation used to represent molecular structures using a compact string of characters \cite{weininger1988smiles}. In SMILES, atoms are represented by letters, where the first letter of an element is uppercase to indicate it is non-aromatic and lowercase to signify it is aromatic. When an element has two letters, the second letter is always lowercase, regardless of aromaticity. Aromaticity can also be denoted by a colon (:) symbol, representing aromatic bonds.

Bond types in SMILES are depicted using specific symbols. The period (.) represents a bond with no connection (isolated atoms), the hyphen (-) indicates a single bond, and the forward-slash (/) and backslash (\textbackslash) are used to represent single bonds adjacent to a double bond, primarily in stereochemical molecules. The equal sign (=) signifies a double bond, the octothorpe (\#) represents a triple bond, and the dollar sign (\$) is used for quadruple bonds.

In stereochemical representations, the "@" symbol is used to denote chirality: a single "@" indicates counterclockwise configuration, while a double "@" denotes clockwise configuration. Additionally, numbers in SMILES denote ring structures by marking the positions where the ring starts and closes. When elements are enclosed in brackets, the number inside indicates the number of atoms involved in the substitution or structure. Isotopes of atoms can be specified by placing a number before the element in square brackets.

\begin{figure}[t]
\vspace{-2mm}
  \centering
  \includegraphics[width=0.5\textwidth]{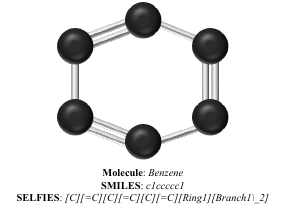}
  \caption{Sample benzene molecule that relates to both the SMILES and SELFIES samples supplied}   \label{fig:benzene}
  \vspace{-6mm}
\end{figure}

Parentheses (()) indicate branches from the main chain, allowing for more complex molecular structures to be represented. The SMILES notation enables efficient and compact molecular representation, facilitating computational processing and analysis in cheminformatics. An example SMILES representation of benzene, shown in Fig.~\ref{fig:benzene}. 


\subsection{Self-Referencing Embedded Strings (SELFIES)}

SELFIES provide a more robust molecular representation for machine learning applications, building on the foundations of SMILES \cite{krenn2020self}. While SMILES offers a simple and interpretable way to encode molecular structures, it has limitations, particularly regarding spatial features. SMILES can represent atoms and bonds but rely on a complex grammar to describe features such as rings and branches, which are not locally represented. This complexity can lead to issues, especially in generative models, where machines often produce syntactically or physically invalid strings.

SELFIES addresses these shortcomings by simplifying the representation of spatial features. Instead of relying on complex grammar to handle rings and branches, SELFIES uses a single symbol to represent these structural elements. The length of the feature, such as the size of a ring or the extent of a branch, is explicitly encoded, ensuring that each SELFIES string is guaranteed to correspond to a valid molecular structure. This approach eliminates the risk of generating invalid or impossible molecules, making SELFIES a powerful tool for generative tasks and other machine-learning applications in cheminformatics. An example SELFIES representation of benzene, shown in Fig.~\ref{fig:benzene}. 


\subsection{Long Short-Term Memory (LSTM)}

Long Short-Term Memory (LSTM) networks \cite{hochreiter1997long} is a specialized type of Recurrent Neural Network (RNN) \cite{elman1990finding} designed to address the limitations of standard RNNs, particularly in handling long-term dependencies in sequential data. Unlike traditional RNNs, which struggle to maintain information over long sequences due to issues such as vanishing or exploding gradients, LSTMs have memory cells that can retain information over extended periods, enabling LSTMs to capture and store temporal information across time steps.

RNNs have an inherent ability to maintain state information, but there is no mechanism in basic RNNs to selectively decide which pieces of information are kept and for how long. This can lead to context saturation, where irrelevant information accumulates, making it difficult for the network to focus on more pertinent data. LSTMs address this problem by introducing specialized gating mechanisms, namely, the input, forget, and output gates, which allow the network to control the flow of information. These gates enable the LSTM to decide when to add new information, remove outdated data, and update its internal state, thus mitigating the saturation issues seen in traditional RNNs and improving performance on tasks requiring long-range dependencies.

\subsection{Quantum Neural Networks (QNN)}
QNNs comprises of three core components: a data encoding circuit, a Variational Quantum Circuit (VQC), and measurement operations. Classical data is first transformed into a quantum state using amplitude, basis, or angle encoding schemes. The VQC, defined by a parameterized ansatz, manipulates this quantum state using single- and two-qubit gates, with entanglement enabling the capture of complex feature correlations \cite{sim2019expressibility}. Finally, measurement operations extract information via quantum observables. During training, VQC parameters are optimized to minimize a task-specific loss function, enabling QNNs to perform machine learning tasks. Integrating quantum circuits with ML frameworks offers promising capabilities for handling high-dimensional and complex data.

\begin{figure}[t]
\vspace{-2mm}
  \centering
  \includegraphics[width=0.48\textwidth]{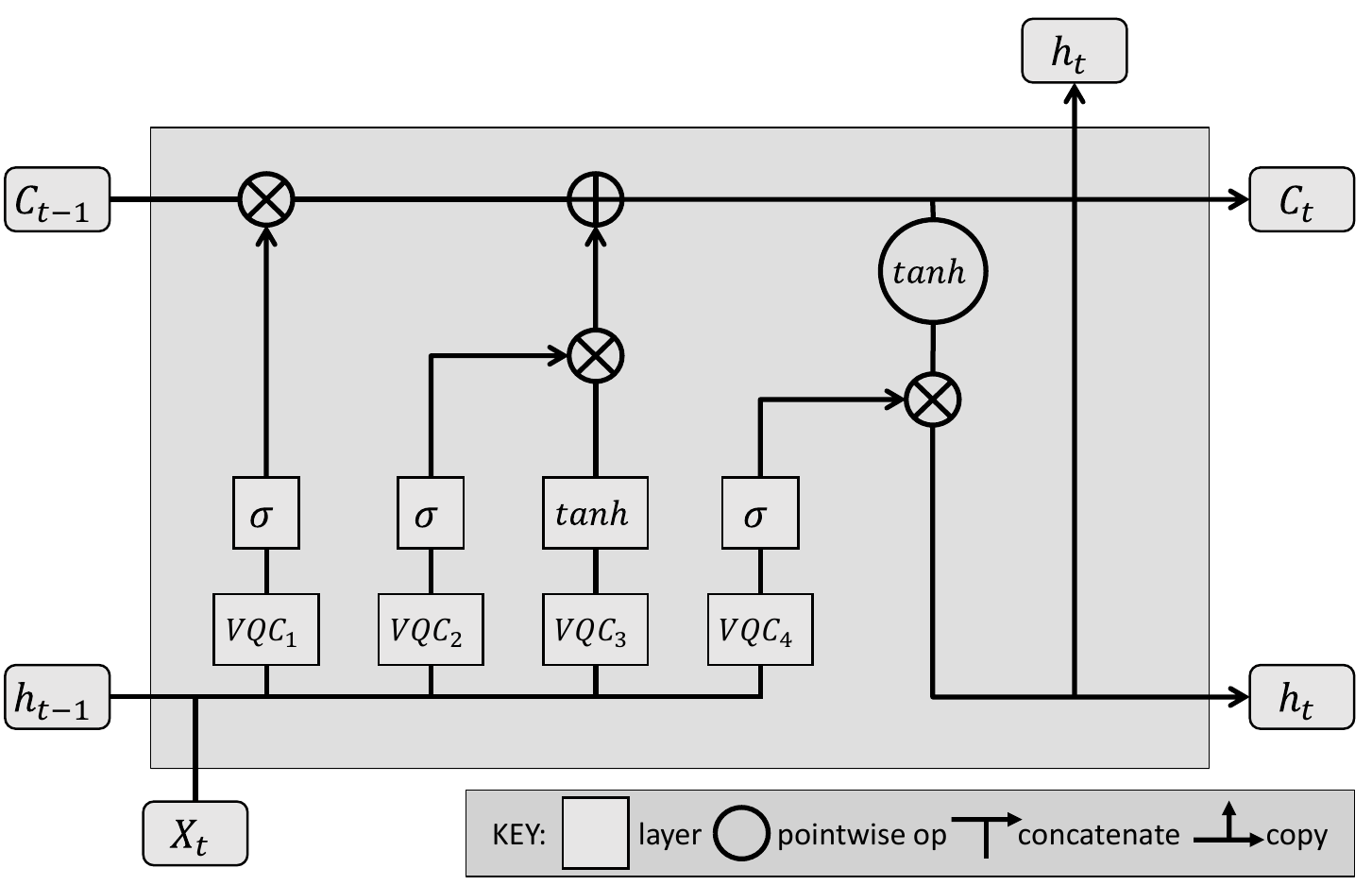}
  \caption{QK-LSTM architecture used for training; ($C_{t-1}, C_t$) represent the cell state, ($X_{t}$) represents the input, ($h_{t-1}, h_t$) represent the output state. The $VQC_1$ wire represents the forget gate, deciding if the input should be added to memory. The ($VQC_2, VQC_3$) wires represent the update gate, updating the cell memory if needed. The $VQC_4$ wire represents the output gate, outputting the result of the QK-LSTM to the rest of the model. }   \label{fig:QLSTM}
  \vspace{-6mm}
\end{figure}

\subsection{Quantum Long Short-Term Memory (QLSTM)}

Despite their promise, prior quantum models such as QNNs struggle to incorporate selective memory mechanisms, key elements of classical machine learning models like LSTM networks. The QLSTM model is proposed to address this, extending the classical LSTM's ability to selectively store and recall contextual information from previous inputs to the quantum realm.
The QLSTM architecture closely mirrors the classical LSTM but rather than directly processing information through traditionally hidden layers, QLSTM passes input data through a VQC. This enables quantum entanglement of the values, which is then measured and processed in the same manner as the classical LSTM's prediction structure \cite{chen2022quantum}. This modification allows QLSTM to exploit quantum phenomena, such as entanglement and superposition, to potentially enhance the model's ability to capture temporal dependencies in sequential data.

\subsection{Quantum Kernel-Based LSTM (QK-LSTM)}

The QK-LSTM \cite{hsu2024quantum} enhances the QLSTM framework by integrating quantum kernels, which map data into higher-dimensional feature spaces \cite{schuld2019quantum}, enabling more expressive representations with fewer trainable quantum parameters. This hybrid approach combines quantum data transformations with classical LSTM mechanisms, improving performance in tasks such as time-series prediction and NLP.

\begin{figure}[t]
\vspace{-2mm}
  \centering
  \includegraphics[width=0.45\textwidth]{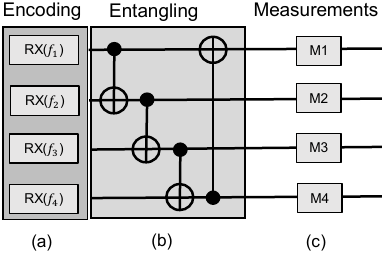}
  \vspace{-2mm}
  \caption{QK-LSTM basic entangler circuit; (a) angle encoding converts classical features ($f_1, f_4, f_3, f_4$) to quantum states, (b) parametric quantum circuit entangles quantum states, (c) qubits measured, and bloated to original higher-dimension space. \cite{bergholm2018pennylane}.} \label{fig:basic_entangler}
\end{figure}

As shown in Fig.\ref{fig:QLSTM}, inputs like $X_t$, $h_t$, and $C_t$ are processed through a sequence of quantum kernels based on entangler circuits (Fig.\ref{fig:basic_entangler}). These kernels enrich the input by leveraging quantum feature maps, followed by a fully connected layer that reduces dimensionality for classical processing.
The enhanced data is passed through standard LSTM components: a forget gate (sigmoid) determines whether to retain the previous context, an input gate (sigmoid and tanh) decides what new information to store, and an output gate (sigmoid) generates predictions and updates the hidden state. This quantum-classical architecture enables QK-LSTM to effectively model complex temporal dependencies while leveraging quantum-enhanced feature extraction for improved sequence learning performance.

\section{Methods \& Results} \label{methods-results}

\subsection{Data Pre-processing} 
The MoleculeNet benchmark dataset \cite{wu2018moleculenet} represents molecules using the SMILES notation. Upon inspection, we observed that not all SMILES strings in the dataset are canonical. Non-canonical SMILES introduce additional complexity due to the inherent syntactic variability of SMILES representations. To mitigate this, all molecular strings are converted to their canonical forms using RDKit, thereby reducing redundancy and improving consistency in downstream learning.

\subsubsection{Augmentation}
Following canonicalization, SMILES augmentation is optionally applied using augmentation techniques proposed by Bjerrum \cite{bjerrum2017smiles}. This technique generates alternative SMILES strings for the same molecule by rearranging the atom and bond sequences while preserving chemical validity. Augmentation increases the diversity of the training data, aiding in model generalization without altering the underlying molecular structure.

\subsubsection{SELFIES Conversion}
To further alleviate SMILES's syntactic complexity and facilitate more robust sequence learning, the canonical and optionally augmented SMILES strings are converted into SELFIES representations \cite{krenn2020self}. SELFIES provide a semantically constrained encoding that ensures the generation of valid molecules. Each SELFIES token, representing atoms or structural motifs such as branches and rings, is mapped to a numerical index based on a predefined vocabulary. This sequence of indices serves as the input to the LSTM-based model.

\subsection{LSTM Implementation}
The pre-processed molecular representations train a recurrent neural network based on the LSTM architecture. Each molecule, represented as a SMILES or SELFIES sequence, is first mapped through an embedding layer, where each token is embedded into a continuous vector space. The embedding dimension corresponds to the size of the molecular vocabulary.

The embedded sequence is then passed into an LSTM network. The LSTM's input and output dimensions match the embedding size, while the hidden state dimensionality is selected empirically to explore different model capacities. The LSTM processes the molecular sequence step-by-step, encoding structural dependencies into its hidden state. The final hidden state is expected to capture the relevant molecular features.

This final representation is fed into a fully connected layer that projects the hidden state to a lower-dimensional space corresponding to the number of output classes—typically, the molecular property to be predicted. This setup lets the model learn end-to-end mappings from molecular structures to target properties.

\begin{figure}[t]
\vspace{-2mm}
  \centering
  \includegraphics[width=.75\linewidth]{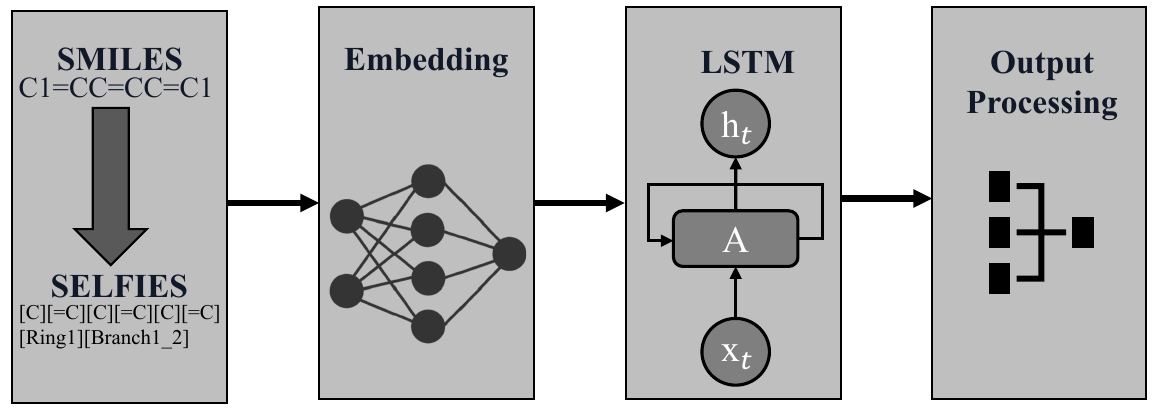}
  \caption{Overview of the LSTM process. }   \label{fig:process_overview}
  \vspace{-6mm}
\end{figure}

\subsection{Results}

Ideally, we would analyze the performance of QK-LSTM across all four variations: SMILES augmented SMILES, SELFIES, and augmented SELFIES. However, the computational cost of running these variations is prohibitive. Therefore, we focus on the two extreme variations, namely SMILES and augmented SELFIES, while comprehensively analyzing the classical model to allow for meaningful abstraction of the results. To evaluate QK-LSTM performance effectively, we employed Optuna \cite{akiba2019optuna} for hyperparameter optimization, randomly selecting four different LSTM and QK-LSTM configurations for training. Each configuration's top three performing models were selected, and the ROC-AUC scores were averaged to derive the final performance metric.

For data augmentation, we followed the recommendation of Bjerrum et al. \cite{bjerrum2023faster}, generating five augmented samples per original data point, where possible. While increasing the number of augmentations could improve the ROC-AUC, it would significantly increase training time. In practice, we generated 20 augmentation samples and selected the five shortest to minimize training time cost.

Following data augmentation, the conversion from SMILES to SELFIES is performed. SMILES are tokenized at the character level, while SELFIES are tokenized at the bracket level, transforming the strings into numerical representations that are subsequently fed into an embedding layer.

The data split is performed randomly following the methodology suggested by MoleculeNet \cite{wu2018moleculenet}, with 80\% of the data allocated for training, 10\% for validation, and 10\% for testing.

Each model is limited to a maximum of 30 epochs to mitigate the impact of excessive training time. Early stopping iss applied if the validation ROC-AUC did not improve over 10 consecutive epochs. We employed a learning rate reduction after five epochs of non-improvement in ROC-AUC to further optimize performance.

Optuna explored the hidden dimension space for the LSTM and QK-LSTM models for hyperparameter optimization. The LSTM's hidden dimension is allowed to range from 32 to 128, with a step size of 16, while the QK-LSTM's hidden dimension varied from 8 to 32, with a step size of 4, resulting in qubit sizes between 3 and 5. The smaller qubit count is chosen to balance reasonable runtime with a realistic comparison of quantum computers' current capabilities against classical implementations.
We used a single model for each side effect classification task to ensure accuracy. The models are trained on an Nvidia GeForce RTX 4090 with an Intel i9-13900K. The final results are averaged across all runs and summarized in Table~\ref{tab:stats}.

\begin{table}[t]
\caption{ROC-AUC Test Score From LSTMs}
\label{tab:stats}
\resizebox{\linewidth}{!}{
\begin{tabular}{@{}c|cc@{}}
\toprule
\textbf{Setup} & \textbf{ROC-AUC} \\
\hline

LSTM SMILES & 0.525 $\pm$ 0.023  \\
LSTM Augmented SMILES & 0.562 $\pm$ 0.004  \\
LSTM SELFIES & 0.507 $\pm$ 0.026  \\
LSTM Augmented SELFIES & 0.556 $\pm$ 0.013 \\ 
QK-LSTM SMILES & 0.524 $\pm$ 0.022 \\
QK-LSTM Augmented SELFIES & 0.555 $\pm$ 0.009 \\ 
\bottomrule
\end{tabular}
}
\end{table}

\subsection{Comparisons}

To understand the performance changes from augmentation, we compare the SMILES to augmented SMILES and SELFIES to augmented SELFIES. From Table ~\ref{tab:stats}, we note a 0.04 increase in the ROC-AUC score (which is outside the standard deviation) when switching between SMILES to augmented SMILES. We get a 0.049 increase in the ROC-AUC score (which is outside the standard deviation as well) for switching between SELFIES and augmented SELFIES. The performance gain for SMILES and SELFIES shows the benefit of increasing the number of available samples to train with, even if they are just variations of the same molecule, as this allows the model to build a more robust understanding of the molecular plane.

To quantify the performance changes from using SELFIES, we compare the SMILES to SELFIES for the standard dataset and the augmented data set. From Table ~\ref{tab:stats}, we note a 0.022 worse performance  (that is within the standard deviation) when switching between SMILES to SELFIES. As for switching between augmented SMILES and augmented SELFIES, we get a 0.006 worse performance change (again within the standard deviation). This shows that we continue to see a non-statistically significant performance change as we increase the size of our dataset. This is problematic for the future usage of SELFIES in molecular property prediction as we continue to build larger and more robust molecular datasets.

We compare SMILES and augmented SELFIES for the LSTM and QK-LSTM to quantify the effect of switching between the classical and quantum domains. Analyzing Table ~\ref{tab:stats}, we note a 0.001 performance loss (that is within the standard deviation) due to switching to the quantum domain for SMILES. As for switching to quantum for augmented SELFIES, we get a 0.001 performance loss (that is within the standard deviation). While quantum cannot outperform the classical domain, we do see the model achieving similar performance gains between SMILES and augmented SELFIES. This is promising for the future of quantum as we continue to advance quantum machine capabilities.

\section{Related Works} \label{related-works}


\subsection{Language Models for Cheminformatics}

\subsubsection{Classical Models}
Prior to Liu et al. \cite{liu2017retrosynthetic}, most works required a structured rules-based approach to predict properties associated with molecules; Liu et al. removed this requirement and achieved similar performance by using LSTMs to make predictions. Unfortunately, it did not result in any significant improvement in state-of-the-art results. 

With the advent of the transformer \cite{vaswani2017attention}, Liu et al.'s proof of concept and the availability of more standardized datasets, such as MoleculeNet \cite{wu2018moleculenet}, many works attempt to leverage the abilities of deep neural networks to attempt to achieve state-of-the-art performance in cheminformatics \cite{irwin2022chemformer, wang2019smiles, taylor2022galactica, zeng2024chatmol, bagal2021molgpt, zhang2021litegem, fang2021chemrl, rong2020self, jin2024large}. While deep learning networks generally show improvement, they come with the cost of exponentially larger datasets and parameter counts.

\subsubsection{Quantum Models}
To overcome the exponential increase of classical parameters, some works have explored the application of quantum machine learning models. Works like Wang et al. \cite{wang2022adverse} attempt to use a QLSTM to identify potential adverse drug reactions. However, the model relies only on Twitter data, ignoring potentially helpful information from chemistry datasets. Works like Beaudoin et al. \cite{beaudoin2022quantum} implement hybrid quantum-classical machine learning models using real cheminformatic datasets to show the potential for hybrid quantum models to perform molecular property prediction, which has recently been confirmed by Zhang et al. \cite{zhang2024quantum}. Unfortunately, none of these works explore how these models can be further improved.

\subsection{Representing Molecules}

Representing molecular structures is pivotal in machine learning-driven drug discovery and cheminformatics. SMILES has become popular among various representations due to its compatibility with text-based deep learning models. However, SMILES is highly complex and limits neural networks' understanding of the input.

Researchers have explored data SMILES enumeration to overcome this limitation, which generates multiple valid SMILES strings for the same molecule by varying the atom order and traversal path. This approach significantly enhances model robustness and generalization. Bjerrum demonstrated that augmenting datasets with randomized SMILES improves the performance of neural networks on molecular property prediction tasks, reducing the risk of overfitting and improving accuracy on unseen molecules \cite{bjerrum2017smiles, arus2019randomized}. However, to our knowledge, no works explore how augmentation can improve SELFIES.

\section{Discussion \& Conclusion} \label{discussion}

While classical models may offer non-statistically significant better performance over near-term hybrid quantum-classical models, we have shown that quantum models are viable candidates for molecular property prediction. Additionally, we have shown that we can leverage classical techniques, such as augmentation, to improve hybrid quantum-classical models further. This will be increasingly important as we continue to scale up our data sizes, as reflected by the results shown by the LSTM.

Interestingly, we see no statistically significant performance gains by converting from SMILES to SELFIES. While SELFIES does guarantee that molecules produced are always real, it does not simplify the learning process for molecules. Guo et al. found that using SELFIES may even result in lower performance for LLMs \cite{guo2023can}, but we did not find this to be the case. However, further evaluation using additional datasets may be warranted for confirmation. The promising news is that the augmentation of SELFIES offers similar performance gains to that of SMILES.

\subsection{Clinical Insights}
Property prediction models allow chemists to perform molecular evaluations prior to physical experimentation. Practical property evaluation can prevent months of wet lab research from being spent on molecules that will not be viable, or may be harmful. Reducing failures realized during synthesis has the potential to significantly reduce the drug-to-market run time, enabling clinical researchers to treat patients for their medical conditions rapidly.

\subsection{Constraints}
Despite LSTM's capabilities and ability to process variable length input with no additional parameters required, such an architecture has drawbacks. LSTM models can scale when dealing with large datasets via batching or even model parallelization, but they do not scale well when considering larger input sequences. Theoretically, LSTM models can process large sequences of information with no problem, but in practice, LSTMs can suffer from vanishing or exploding gradients just like classic RNNs, causing them to "forget" important information. Even if we could implement the perfect memory model, the LSTM would still suffers from long run times where each addition to the sequence increases the run time due to the sequential nature of recurrence. One possible method to mitigate the long run time would be chunking, where the sequences are partitioned into smaller processable pieces. Unfortunately, this is unreliable, as sometimes vital state information may be separated from chunks, causing inaccurate results.

\subsection{Ethical Statement}
While machine learning models can help identify potential molecular properties, they have flaws. Even if machine learning models can accurately identify all molecular properties of the datasets they are trained with, they depend entirely on previous human discoveries. The properties associated with molecules are subject to our current understanding of them. For neuvo molecular designs we simply do not have enough understanding of chemistry to be certain of our predictions. The datasets used for property identification are subject to flawed understandings of chemistry and even political choices. For example, the NIH only classifies drugs as toxic to the liver after successfully ruling out other potential causes \footnote{https://www.ncbi.nlm.nih.gov/books/NBK548049/}. Therefore, machine learning models should only be used for the preliminary evaluation of molecules and not as the only form of molecular evaluation.

\section*{Acknowledgment}

The work is supported in parts by NSF (CNS-1722557 and CNS-2129675) and gifts from Intel.

\bibliographystyle{IEEEtran}

\bibliography{test}

\begin{thebibliography}{10}
\providecommand{\url}[1]{#1}
\csname url@samestyle\endcsname
\providecommand{\newblock}{\relax}
\providecommand{\bibinfo}[2]{#2}
\providecommand{\BIBentrySTDinterwordspacing}{\spaceskip=0pt\relax}
\providecommand{\BIBentryALTinterwordstretchfactor}{4}
\providecommand{\BIBentryALTinterwordspacing}{\spaceskip=\fontdimen2\font plus
\BIBentryALTinterwordstretchfactor\fontdimen3\font minus \fontdimen4\font\relax}
\providecommand{\BIBforeignlanguage}[2]{{%
\expandafter\ifx\csname l@#1\endcsname\relax
\typeout{** WARNING: IEEEtran.bst: No hyphenation pattern has been}%
\typeout{** loaded for the language `#1'. Using the pattern for}%
\typeout{** the default language instead.}%
\else
\language=\csname l@#1\endcsname
\fi
#2}}
\providecommand{\BIBdecl}{\relax}
\BIBdecl

\bibitem{yang2019analyzing}
K.~Yang \emph{et~al.}, ``Analyzing learned molecular representations for property prediction,'' \emph{Journal of chemical information and modeling}, vol.~59, no.~8, pp. 3370--3388, 2019.

\bibitem{wieder2020compact}
O.~Wieder \emph{et~al.}, ``A compact review of molecular property prediction with graph neural networks,'' \emph{Drug Discovery Today: Technologies}, vol.~37, pp. 1--12, 2020.

\bibitem{lengauer2004novel}
T.~Lengauer \emph{et~al.}, ``Novel technologies for virtual screening,'' \emph{Drug discovery today}, vol.~9, no.~1, pp. 27--34, 2004.

\bibitem{merkwirth2005automatic}
C.~Merkwirth \emph{et~al.}, ``Automatic generation of complementary descriptors with molecular graph networks,'' \emph{Journal of chemical information and modeling}, vol.~45, no.~5, pp. 1159--1168, 2005.

\bibitem{hu2019strategies}
W.~Hu \emph{et~al.}, ``Strategies for pre-training graph neural networks,'' \emph{arXiv preprint arXiv:1905.12265}, 2019.

\bibitem{wu2020comprehensive}
Z.~Wu \emph{et~al.}, ``A comprehensive survey on graph neural networks,'' \emph{IEEE transactions on neural networks and learning systems}, vol.~32, no.~1, pp. 4--24, 2020.

\bibitem{mayr2018large}
Mayr \emph{et~al.}, ``Large-scale comparison of machine learning methods for drug target prediction on chembl,'' \emph{Chemical science}, vol.~9, no.~24, pp. 5441--5451, 2018.

\bibitem{taylor2022galactica}
R.~Taylor \emph{et~al.}, ``Galactica: A large language model for science,'' \emph{arXiv preprint arXiv:2211.09085}, 2022.

\bibitem{m2024augmenting}
A.~M.~Bran, S.~Cox, O.~Schilter, C.~Baldassari, A.~D. White, and P.~Schwaller, ``Augmenting large language models with chemistry tools,'' \emph{Nature Machine Intelligence}, vol.~6, no.~5, pp. 525--535, 2024.

\bibitem{jin2024large}
B.~Jin, G.~Liu, C.~Han, M.~Jiang, H.~Ji, and J.~Han, ``Large language models on graphs: A comprehensive survey,'' \emph{IEEE Transactions on Knowledge and Data Engineering}, 2024.

\bibitem{wang2019smiles}
Wang \emph{et~al.}, ``Smiles-bert: large scale unsupervised pre-training for molecular property prediction,'' in \emph{Proceedings of the 10th ACM international conference on bioinformatics, computational biology and health informatics}, 2019, pp. 429--436.

\bibitem{irwin2022chemformer}
R.~Irwin, S.~Dimitriadis, J.~He, and E.~J. Bjerrum, ``Chemformer: a pre-trained transformer for computational chemistry,'' \emph{Machine Learning: Science and Technology}, vol.~3, no.~1, p. 015022, 2022.

\bibitem{liu2024git}
P.~Liu, Y.~Ren, J.~Tao, and Z.~Ren, ``Git-mol: A multi-modal large language model for molecular science with graph, image, and text,'' \emph{Computers in biology and medicine}, vol. 171, p. 108073, 2024.

\bibitem{tetko2019augmentation}
I.~V. Tetko, P.~Karpov, E.~Bruno, T.~B. Kimber, and G.~Godin, ``Augmentation is what you need!'' in \emph{International Conference on Artificial Neural Networks}.\hskip 1em plus 0.5em minus 0.4em\relax Springer, 2019, pp. 831--835.

\bibitem{bjerrum2017smiles}
E.~J. Bjerrum, ``Smiles enumeration as data augmentation for neural network modeling of molecules,'' \emph{arXiv preprint arXiv:1703.07076}, 2017.

\bibitem{beaudoin2022quantum}
C.~Beaudoin, S.~Kundu, R.~O. Topaloglu, and S.~Ghosh, ``Quantum machine learning for material synthesis and hardware security,'' in \emph{Proceedings of the 41st IEEE/ACM International Conference on Computer-Aided Design}, 2022, pp. 1--7.

\bibitem{mao2024advancing}
Q.~Mao, Z.~Liu, C.~Liu, Z.~Li, and J.~Sun, ``Advancing graph representation learning with large language models: A comprehensive survey of techniques,'' \emph{arXiv preprint arXiv:2402.05952}, 2024.

\bibitem{ozcceliknovo}
R.~{\"O}z{\c{c}}elik and F.~Grisoni, ``De novo drug design by chemical language modeling,'' in \emph{An Introduction to Generative Drug Discovery}.\hskip 1em plus 0.5em minus 0.4em\relax CRC Press, pp. 45--66.

\bibitem{wu2018moleculenet}
Wu \emph{et~al.}, ``Moleculenet: a benchmark for molecular machine learning,'' \emph{Chemical science}, vol.~9, no.~2, pp. 513--530, 2018.

\bibitem{kuhn2016sider}
M.~Kuhn \emph{et~al.}, ``The sider database of drugs and side effects,'' \emph{Nucleic acids research}, vol.~44, no.~D1, pp. D1075--D1079, 2016.

\bibitem{fawcett2006introduction}
T.~Fawcett, ``An introduction to roc analysis,'' \emph{Pattern recognition letters}, vol.~27, no.~8, pp. 861--874, 2006.

\bibitem{weininger1988smiles}
D.~Weininger, ``Smiles, a chemical language and information system. 1. introduction to methodology and encoding rules,'' \emph{Journal of chemical information and computer sciences}, vol.~28, no.~1, pp. 31--36, 1988.

\bibitem{krenn2020self}
M.~Krenn \emph{et~al.}, ``Self-referencing embedded strings (selfies): A 100\% robust molecular string representation,'' \emph{Machine Learning: Science and Technology}, vol.~1, no.~4, p. 045024, 2020.

\bibitem{hochreiter1997long}
Hochreiter \emph{et~al.}, ``Long short-term memory,'' \emph{Neural computation}, vol.~9, no.~8, pp. 1735--1780, 1997.

\bibitem{elman1990finding}
J.~L. Elman, ``Finding structure in time,'' \emph{Cognitive science}, vol.~14, no.~2, pp. 179--211, 1990.

\bibitem{sim2019expressibility}
S.~Sim, P.~D. Johnson, and A.~Aspuru-Guzik, ``Expressibility and entangling capability of parameterized quantum circuits for hybrid quantum-classical algorithms,'' \emph{Advanced Quantum Technologies}, vol.~2, no.~12, p. 1900070, 2019.

\bibitem{chen2022quantum}
S.~Y.-C. Chen, S.~Yoo, and Y.-L.~L. Fang, ``Quantum long short-term memory,'' in \emph{ICASSP 2022-2022 IEEE International Conference on Acoustics, Speech and Signal Processing (ICASSP)}.\hskip 1em plus 0.5em minus 0.4em\relax IEEE, 2022, pp. 8622--8626.

\bibitem{hsu2024quantum}
Y.-C. Hsu, T.-Y. Li, and K.-C. Chen, ``Quantum kernel-based long short-term memory,'' \emph{arXiv preprint arXiv:2411.13225}, 2024.

\bibitem{schuld2019quantum}
M.~Schuld and N.~Killoran, ``Quantum machine learning in feature hilbert spaces,'' \emph{Physical review letters}, vol. 122, no.~4, p. 040504, 2019.

\bibitem{bergholm2018pennylane}
V.~Bergholm, J.~Izaac, M.~Schuld, C.~Gogolin, M.~S. Alam, S.~Ahmed, J.~M. Arrazola, C.~Blank, A.~Delgado, S.~Jahangiri \emph{et~al.}, ``Pennylane: Automatic differentiation of hybrid quantum-classical computations,'' \emph{arXiv preprint arXiv:1811.04968}, 2018.

\bibitem{akiba2019optuna}
T.~Akiba, S.~Sano, T.~Yanase, T.~Ohta, and M.~Koyama, ``{O}ptuna: A next-generation hyperparameter optimization framework,'' in \emph{The 25th ACM SIGKDD International Conference on Knowledge Discovery \& Data Mining}, 2019, pp. 2623--2631.

\bibitem{bjerrum2023faster}
E.~J. Bjerrum, C.~Margreitter, T.~Blaschke, S.~Kolarova, and R.~L.-R. de~Castro, ``Faster and more diverse de novo molecular optimization with double-loop reinforcement learning using augmented smiles,'' \emph{Journal of Computer-Aided Molecular Design}, vol.~37, no.~8, pp. 373--394, 2023.

\bibitem{liu2017retrosynthetic}
B.~Liu, B.~Ramsundar, P.~Kawthekar, J.~Shi, J.~Gomes, Q.~Luu~Nguyen, S.~Ho, J.~Sloane, P.~Wender, and V.~Pande, ``Retrosynthetic reaction prediction using neural sequence-to-sequence models,'' \emph{ACS central science}, vol.~3, no.~10, pp. 1103--1113, 2017.

\bibitem{vaswani2017attention}
Vaswani \emph{et~al.}, ``Attention is all you need,'' \emph{Advances in neural information processing systems}, vol.~30, 2017.

\bibitem{zeng2024chatmol}
Z.~Zeng, B.~Yin, S.~Wang, J.~Liu, C.~Yang, H.~Yao, X.~Sun, M.~Sun, G.~Xie, and Z.~Liu, ``Chatmol: interactive molecular discovery with natural language,'' \emph{Bioinformatics}, vol.~40, no.~9, p. btae534, 2024.

\bibitem{bagal2021molgpt}
V.~Bagal, R.~Aggarwal, P.~Vinod, and U.~D. Priyakumar, ``Molgpt: molecular generation using a transformer-decoder model,'' \emph{Journal of chemical information and modeling}, vol.~62, no.~9, pp. 2064--2076, 2021.

\bibitem{zhang2021litegem}
Zhang \emph{et~al.}, ``Litegem: Lite geometry enhanced molecular representation learning for quantum property prediction,'' \emph{arXiv preprint arXiv:2106.14494}, 2021.

\bibitem{fang2021chemrl}
Fang \emph{et~al.}, ``Chemrl-gem: Geometry enhanced molecular representation learning for property prediction,'' \emph{arXiv preprint arXiv:2106.06130}, 2021.

\bibitem{rong2020self}
Rong \emph{et~al.}, ``Self-supervised graph transformer on large-scale molecular data,'' \emph{Advances in Neural Information Processing Systems}, vol.~33, pp. 12\,559--12\,571, 2020.

\bibitem{wang2022adverse}
X.~Wang, X.~Wang, and S.~Zhang, ``Adverse drug reaction detection from social media based on quantum bi-lstm with attention,'' \emph{IEEE Access}, vol.~11, pp. 16\,194--16\,202, 2022.

\bibitem{zhang2024quantum}
L.~Zhang, Y.~Xu, M.~Wu, L.~Wang, and H.~Xu, ``Quantum long short-term memory for drug discovery,'' \emph{arXiv preprint arXiv:2407.19852}, 2024.

\bibitem{arus2019randomized}
J.~Ar{\'u}s-Pous, S.~V. Johansson, O.~Prykhodko, E.~J. Bjerrum, C.~Tyrchan, J.-L. Reymond, H.~Chen, and O.~Engkvist, ``Randomized smiles strings improve the quality of molecular generative models,'' \emph{Journal of cheminformatics}, vol.~11, pp. 1--13, 2019.

\bibitem{guo2023can}
T.~Guo, B.~Nan, Z.~Liang, Z.~Guo, N.~Chawla, O.~Wiest, X.~Zhang \emph{et~al.}, ``What can large language models do in chemistry? a comprehensive benchmark on eight tasks,'' \emph{Advances in Neural Information Processing Systems}, vol.~36, pp. 59\,662--59\,688, 2023.

\end{thebibliography}

\end{document}